\documentclass[10pt,journal]{IEEEtran}
\ifCLASSINFOpdf
\else
\fi
\usepackage{tabularx}
\usepackage{array}
\usepackage{pifont}
\usepackage{graphicx}
\usepackage{caption}
\usepackage{amsfonts}
\usepackage{amssymb}
\usepackage{mathtools}
\usepackage{tikz}
\usepackage{bbding}
\usepackage{tikz}
\usepackage{tabularx}
\usepackage{array}
\usepackage{amsmath}
\usepackage{pifont}
\usepackage{graphicx}
\usepackage{mathtools}
\usepackage{tikz}
\usepackage{bbding}
\usepackage{tikz}
\usepackage{cite}
\usepackage{hyperref}
\usepackage{ragged2e}
\usepackage{hyperref}
\usepackage{cite}
\usepackage{epsfig}
\usepackage{epstopdf}
\usepackage{tabularx}
\usepackage{array}
\usepackage{amsmath}
\usepackage{slashbox}
\usepackage{pifont}
\usepackage{graphicx}
\usepackage{amsfonts}
\usepackage{amssymb}
\usepackage{mathtools}
\usepackage{amsmath}
\usepackage{tikz}
\usepackage{bbding}
\usepackage{tikz}
\usepackage{cite}
\usepackage{hyperref}
\usepackage{subcaption}
\usepackage{csquotes}

\usepackage{caption}
\DeclareCaptionFormat{citation}{%
  \ifx\captioncitation\relax\relax\else
    \captioncitation\par
  \fi
  #1#2#3\par}
\newcommand*\setcaptioncitation[1]{\def\captioncitation{\textit{Source:}~#1}}
\let\captioncitation\relax
\usepackage[compact]{titlesec}
    \titlespacing{\section}{0pt}{2ex}{1ex}
    \titlespacing{\subsection}{0pt}{1ex}{0ex}
    \titlespacing{\subsubsection}{0pt}{0.5ex}{0ex}

\usepackage{amsthm}

\begin{document}
%
%
\title{ An approach to implement Reinforcement Learning for Heterogeneous Vehicular Networks}
%
%
\author{
        Bhavya Peshavaria,
        Sagar Kavaiya,
        Dhaval K. Patel,~\IEEEmembership{Member,~IEEE}
       
        
        \thanks{Bhavya Peshavaria, Sagar Kavaiya and Dhaval K Patel are with School of Engineering and Applied Science, Ahmedabad University, India. (Email:{bhavya.p@ahduni.edu.in,sagar.k@ahduni.edu.in, dhaval.patel@ahduni.edu.in}).

}}
\maketitle
\begin{abstract}
This paper presents the extension of the idea of spectrum sharing in the vehicular networks towards the Heterogeneous Vehicular Network(HetVNET) based on multi-agent reinforcement learning. Here, the multiple vehicle-to-vehicle(V2V) links reuse the spectrum of other vehicle-to-interface(V2I) and also those of other networks. The fast-changing environment in vehicular networks limits the idea of centralizing the CSI and allocate the channels. So, the idea of implementing ML-based methods is used here so that it can be implemented in a distributed manner in all vehicles. Here each On-Board Unit(OBU) can sense the signals in the channel and based on that information runs the RL to decide which channel to autonomously take up. Here, each V2V link will be an agent in MARL. The idea is to train the RL model in such a way that these agents will collaborate rather than compete.
\end{abstract}


%
\IEEEpeerreviewmaketitle

\section{Introduction}
%
%
%
%
Vehicular communication had been introduced first time in 3GPP Release 12 for communications with devices with mobility named D2D communication. Upon its evolution towards release 14 and later such a communication link was termed as vehicle-to-everything(V2X). Vehicular communication has achieved a high amount of attention from research scholars and companies as it is a key enabler of Intelligent Transportation Systems (ITS), smart cities, and autonomous driving. In the recent 3GPP release 16, the concept of vehicular communication is broadened to improve road safety by letting the On-Board Units(OBUs) of cars communicate with each other. Such communications include disseminating safety messages, traffic hazards ahead and other sensor-based information picked up by the system. These communications are meant to improve road safety, the comfort of transport, traffic congestion problems and enhance situational awareness. However, due to the high mobility of the vehicles, it is difficult to establish a reliable communication channel between two vehicles to satisfy the necessity of lower latency.

Besides, the use cases of the V2X have been extended to incorporate autonomous driving. According to the SAE International’s Standard J3016 [1], vehicles are classified into six levels of autonomy. Of which, the vehicles with the autonomy levels 0 and 1 are manually driven with no technological assistance. However, the vehicles of levels 3 and 4 have partial and conditional autonomy; here, vehicular communication could be implemented to let the drivers know of the information about other vehicles. Those of levels 4 and 5 have a high level of autonomy and no need for human interference. Such autonomous vehicles need to constantly communicate among those in the vicinity to deliver messages about their actions so that other vehicles can decide their actions. Such a scenario needs highly reliable and low latency communication.\\

The focus of this paper will be on optimizing the communication in the vehicle-to-vehicle(V2V) links. These links have one crucial necessity of having low latency as they are responsible for disseminating the safety messages for other vehicles and user equipment. \\

The paper is organized as follows. In Section I(A), the background of the concepts involved in this paper is given. It includes a brief of different works of literature in heterogeneous vehicular networks. In Section II, we present the key takeaways and understanding of the work done throughout this duration. Section III describes the future work which will be carried out in the following months. It has plans on how to connect the dots of the concepts discussed in this paper to bring out innovation. \\

\subsection{ \textcolor{black}{Background}}

\begin{figure}[htp]
    \centering
    \includegraphics[width=10cm]{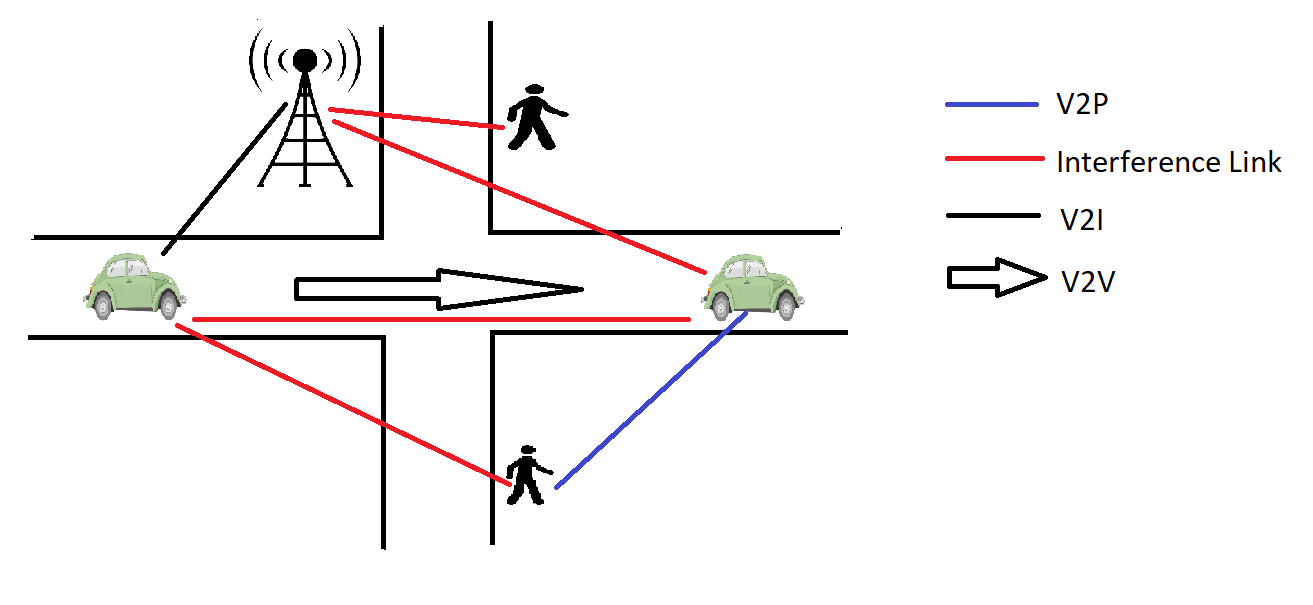}
    \caption{A visual representation of the communication links in the 3GPP Release 16 for C-V2X communication}
    \label{fig:v2xlinks}
\end{figure}

According to 3GPP release 16, the following types of communication are included in vehicular communication:
Vehicle-to-vehicle (V2V) :
This type of communication will be utilized for the communication among vehicles. The main use of it would be to transmit the sensor data, traffic hazards, vehicle breakdown and other road safety data. Use cases of this can vary from platooning, sending warning messages of vehicles out of site and event-driven safety information.
Vehicle-to-infrastructure (V2I)
This type of link will serve the purpose of providing a reliable and low-latency data connection to the internet for the user devices inside the vehicle. Such links will mainly be used for entertainment purposes in vehicles.
Vehicle-to-pedestrian (V2P) 
This link is very similar to V2V links. The idea behind this is to disseminate the warning messages for the pedestrians of any possible accidents. Use cases of this could be to warn pedestrians of any vehicular issuers like brake failure so that they have enough time to react. Besides, it can be used to warn pedestrians about the vehicles out of sight.
In [3], the idea of spectrum sharing in cellular networks has been discussed. The goal in [3] is to utilize unused and available bands of the licensed cellular network for unlicensed V2V communication so as to improve the throughput of dissemination of the safety messages while not causing any disturbance in the currently occupied cellular bands. The concept used there was to use a cellular network(Uu) for the V2I links to provide a reliable and fast internet connection to the User Equipment(UE), whereas use sidelink(PC5) for the V2V communication. This sidelink was according to 3GPP Release 15, mode 4 where the UE can sense the channels in the sensing window and then based on the received CSI, autonomously select a band for the V2V link. A multi-agent reinforcement learning(MARL) framework is used on the UE to autonomously pick the most efficient selection scheme in the sensed conditions. This way, by implementing spectrum sharing, the available spectrum can be better utilized without needing to clear out the bands for such ad-hoc communications.

In [5], a cascaded Hungarian channel allocation algorithm has been described for non-orthogonal multiple access (NOMA) based heterogeneous vehicular networks. A chance-constrained throughput optimization problem has been formulated to combat the imperfect CSI which arises due to high mobility in vehicular networks.

[6] has utilized the white space spectrum in TV networks as an approach to implementing Heterogeneous vehicular networks. They chose a non-cooperative game theory approach with correlated equilibrium for the allocation of resources. Here, the idea is to allow the macro-cell base station to allow sharing available unused resources to UE without causing much interference to macro-cell performance. 

[7] has presented QoS based resource allocation scheme for an SDN(Software-defined network) by sharing the LTE and Wi-Fi networks. A heuristic solution was used for RA to be used in central as well as hybrid SDN. The simulation consisted of 20 vehicles, one remote server, an eNodeB and 3 Wi-Fi AP(Access Points). The proposed schemes were then evaluated using NS-3.

An approach about HetVNET for LTE and DSRC has been proposed in [8]. The approach first tries to allocate all the LTE resources optimally and that way reduce the channel congestion in DSRC.\\

\section{\textcolor{black}{Contribution}}

\medskip

\subsection{Study of MARL}

\begin{figure}[htp]
    \centering
    \includegraphics[width=9cm]{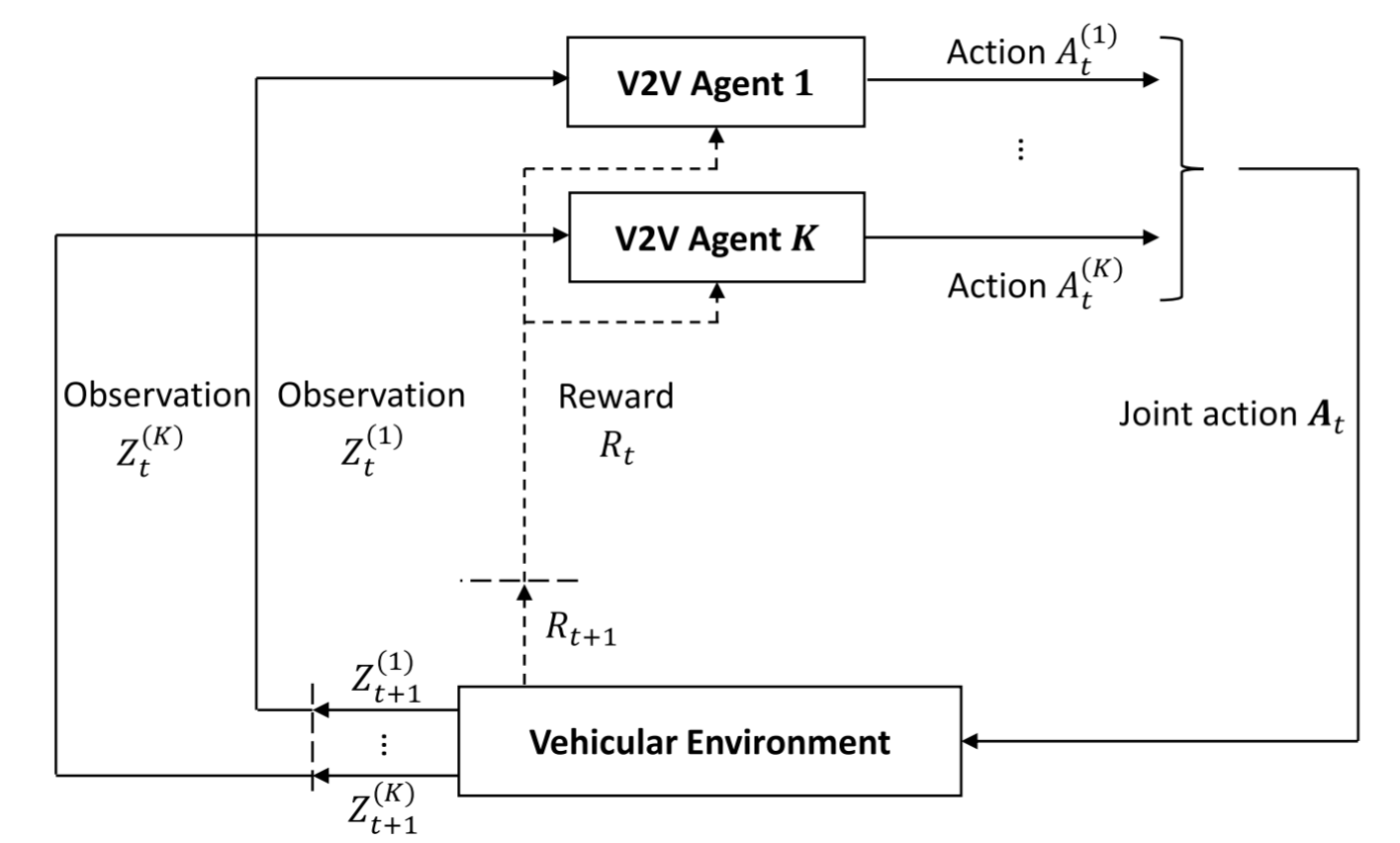}
    \caption{Logical flow of MARL as discussed in [3]}
    \label{fig:galaxy}
     \setcaptioncitation{Source: [3], fig 2}
\end{figure}

\medskip

The fig. 2 is the representation of the flow of RL implemented in [1]. In this scenario, the V2I links are already assigned, the task is to assign V2V links. Here, each V2V link is an agent in the RL framework.

The simulation starts with all the k agents(V2V links) getting their respective observations from the environment. The observation function Zt(k) represents the observation of the kth agent at time t. Once all the agents have received their observations, they take independent decisions on which available links to use and then take corresponding actions. Here, all the agents take action simultaneously. The action function At(k) represents the action of the kth agent at time t. Once the actions are taken, the joint action At goes to the environment to simulate the data transfer. Once the simulations are finished, the rewards are given to all the agents according to their actions. And then the simulation goes to the next iteration at time t+1. However, it waits till all the agents have updated their Q network based on the result at the given time t. Once all the agents are ready, they observe the environment again from the observation function. This was one iteration.\\

\subsection{\textcolor{black}{Reward Function}}
The reward function is a key element of the RL framework. This function takes the simulated result of the actions of agents as input for processing. It then rates this input according to the metric coded in it to rate the decisions of the agents. So, if the actions of the agent lead it closer to the defined goal, then it will give more reward and vice versa. This function basically takes the simulation towards the necessary goal.

In [1] reward function was needed to be carefully designed. As all the agents are working in the same environment, the actions of one will affect the other. Say the reward function is kept simply as each agent gets the reward according to their successful transmission. In this scenario, the agents will evolve to compete rather than collaborate. Each agent will try to hog as many spectral resources as it can to get more rewards for itself. Such a reward function will not give us the model we need.

So, in order to establish collaborative efforts among the agents, the reward function is designed in a different way. Here, the scheme is that all the agents will get a reward only if each of the agents has successfully completed its transmission; if one agent fails then none would get the reward. In this way, the agents will now try to collaborate with each other to optimize the resource allocation such that they collectively achieve the maximal efficiency they can in the given situation.

In this way, the reward function is defined in [1] to achieve collaboration instead of competition.\\

\subsection{Reproduced Results}
The following results are reproduced from [1]. 
Note: Due to less computation power, there is less accuracy. This leads to the graph differing from the original paper.\\

\subsubsection{Figure 3 of [3]}

\begin{figure}[htp]
    \centering
    \includegraphics[width=10cm]{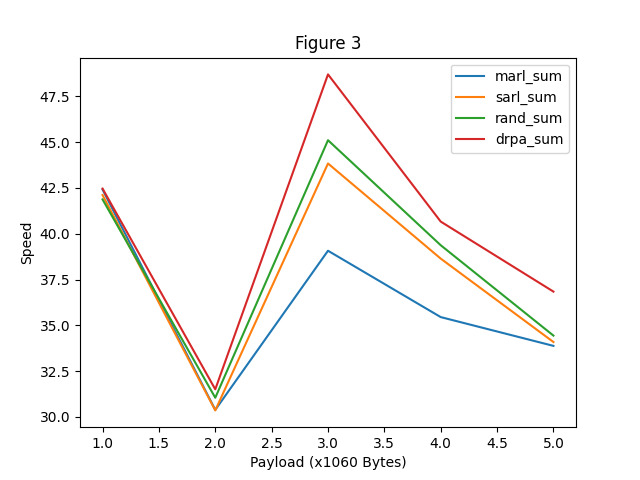}
    \caption {Sum  capacity  performance of  V2I  links with varying payload of V2V}
    \label{fig:galaxy}
\end{figure}

In fig. 3  the blue line(marl\_sum) shows the sum capacity of V2I when RA of V2V was done with the help of the proposed MARL and the orange line (sarl\_sum) shows the sum of the sum capacity of V2I when RA of V2V was done with SARL(Single agent reinforcement learning).

The fig. 3  shows the graph of sum capacity performance of V2I links with varying V2V payload sizes. As we can observe from the graph that the general trend of the sum of speeds of V2I links is decreasing with the increase in the payload size of V2V links. This can be explained by the fact that V2V starts occupying more bandwidth and hence affecting the performance of V2I. The main goal in [1] was to allocate the resources optimally so that the sum of V2I speed remains maximal while sharing the spectrum with V2V links.\\

\subsubsection{Figure 4 of [3]}

\begin{figure}[htp]
    \centering
    \includegraphics[width=10cm]{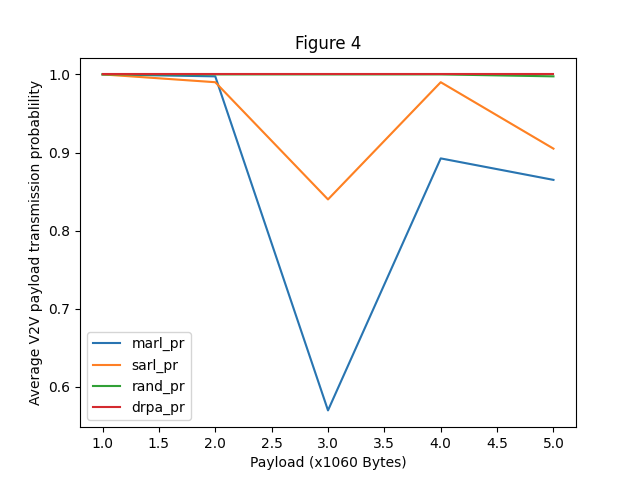}
    \caption{V2V payload transmission success probability with varying payload of V2V}
    \label{fig:galaxy}
\end{figure}

In fig. 4 the blue line(marl\_sum) shows the V2V payload transmission success probability with varying payload sizes when done with the help of the proposed MARL and the orange line (sarl\_sum) shows the V2V payload transmission success probability with varying payload sizes when RA of V2V was done with SARL(Single agent reinforcement learning).

The fig. 4  shows the V2V payload transmission success probability with varying payload sizes. It can be observed from the graph trend of all the lines that with increasing transmission payload, the probability of successful transmission decreases. This can be explained as the loss of data packets over the transmission. The payload which is to be passed is broken down into smaller chunks and then sent over the network. In this, if one packet gets lost, then the whole payload is said to fail. So, the larger the payload, the more the packets and hence the higher chances of transmission failure. This in turn decreases the successful payload transmission as the payload size increases.\\

\section{Future Work}
With the increasing number of devices working on wireless communication technology, it is possible to have a high UE density in some regions. In such cases, the latency and packet drop may increase in that particular network. Such conditions are not favourable for V2V transmission as they have time crucial data which needs to be sent with high reliability and low latency. In such cases, it is beneficial to switch towards other networks. 

As discussed in the background section, there has been a lot of literature around using heterogeneous vehicular networks. By combining the works in [5], [6], [7] and [8] a HetVNET of the white TV spectrum, DSRC, Wi-Fi AP and LTE/NR can be achieved. Such a large heterogeneous network has been discussed in [4].

To proceed further, the selection of which network to use at any given instance needs to be decided in real-time. There have been some opportunistic and greedy algorithms for such tasks but they cannot be malleable to all situations and environments.

The concept of allocating the spectrum according to the environment had been discussed in [3]. There, the concept was to allocate mobile network resources to the V2V links dynamically. Now, that concept can be extended to work with the HetVNET. In this model, the OBU will sense the channels of all the networks enabled (i.e. white TV spectrum, DSRC, Wi-Fi AP and 5G LTE or NR)and then based on the environment and the CSI will decide which band of which network to choose for communication.\\

\section{End Goal}
As discussed in the previous section about future work, the idea is to extend the concept of the implementation of MARL for efficient allocation of resources to include all the other possible networks of HetVNET. This way there will be a diverse range of spectral resources available for communication and the agents can choose which are the most efficient links to communicate over in the given scenario.

For example, if the communicating vehicles are going at similar speeds in the same direction then they would choose DSRC over cellular as they are going to stay in range for long enough for successful transmission and the latency of DSRC is lower than Uu. Similarly, the agents can choose other networks based on their pros and cons.

This implementation of HetVNET will also act as a load balancing mechanism for the traffic in all the networks. If a network would get too crowded in terms of resource requests, then the devices with this technology enabled could easily switch over to the other network to offload the burden on the network while making sure their transmission is not hindered by the network traffic.\\

\section{Conclusion}
This paper has discussed the idea of combining HetVNET and MARL for resource allocation on a superficial level as an in-depth study of this concept is out of scope here. The main aim here is to impart this concept to other scholars for the implementation and research of this in further studies. There are more challenges to the implementation of such a technology at the present time. Here each OBU is assumed to be having the necessary hardware resource to process the whole MARL based framework in real-time. Such hardware could be tough to fit in vehicles and consume more energy for performing the necessary processes. However, when implemented, this concept would let the autonomous and semi-autonomous vehicles communicate with each other in a more reliable way. \\

\section{Reference}
1. Taxonomy and Definitions for Terms Related to On-Road Motor Vehicle Automated Driving Systems, SAE Standard J3016, 2016.\\

2. M. Noor-A-Rahim, Z. Liu, H. Lee, G. G. M. N. Ali, D. Pesch and P. Xiao, "A Survey on Resource Allocation in Vehicular Networks," in IEEE Transactions on Intelligent Transportation Systems, doi: 10.1109/TITS.2020.3019322.\\
\\3. Liang, Le, Hao Ye, and Geoffrey Ye Li. "Spectrum sharing in vehicular networks based on multi-agent reinforcement learning." IEEE Journal 3. on Selected Areas in Communications 37.10 (2019): 2282-2292. \\
\\4. H. Peng, Le Liang, X. Shen and G. Y. Li, "Vehicular Communications: A Network Layer Perspective," in IEEE Transactions on Vehicular Technology, vol. 68, no. 2, pp. 1064-1078, Feb. 2019, doi: 10.1109/TVT.2018.2833427.\\
\\5. S. Guo and X. Zhou, "Robust Resource Allocation With Imperfect Channel Estimation in NOMA-Based Heterogeneous Vehicular Networks," in IEEE Transactions on Communications, vol. 67, no. 3, pp. 2321-2332, March 2019, doi: 10.1109/TCOMM.2018.2885999.\\
\\6. Z. Xiao et al., "Spectrum Resource Sharing in Heterogeneous Vehicular Networks: A Noncooperative Game-Theoretic Approach With Correlated Equilibrium," in IEEE Transactions on Vehicular Technology, vol. 67, no. 10, pp. 9449-9458, Oct. 2018, doi: 10.1109/TVT.2018.2855683.\\
\\7. W. Huang, L. Ding, D. Meng, J. Hwang, Y. Xu and W. Zhang, "QoE-Based Resource Allocation for Heterogeneous Multi-Radio Communication in Software-Defined Vehicle Networks," in IEEE Access, vol. 6, pp. 3387-3399, 2018, doi: 10.1109/ACCESS.2018.2800036.\\
\\8. X. Cao, L. Liu, Y. Cheng, L. X. Cai, and C. Sun, “On optimal Device-to-Device resource allocation for minimizing end-to-end delay in VANETs,” IEEE Trans. Veh. Technol., vol. 65, no. 10, pp. 7905–7916, Oct. 2016.

\end{document}